\documentclass[10pt,onecolumn,twoside]{IEEEtran}

\usepackage{amsmath}
\usepackage{ifpdf}
\usepackage{array}
\usepackage{url}
\usepackage{graphicx}

\ifpdf
\DeclareGraphicsExtensions{.pdf}
\else
\DeclareGraphicsExtensions{.eps}
\fi

\graphicspath{{./}}

\usepackage{fancyhdr}
\begin{document}

\title{Using a Kernel Adatron for\\
Object Classification with RCS Data}

\author{Marten~F.~Byl, James~T.~Demers, and Edward~A.~Rietman\thanks{Marten~F.~Byl, James T. Demers, and Edward A. Rietman are with Physical Sciences Inc., 20 New England Business Center, Andover, MA 01810-1077.  Tel:~(978)~689-0003.  Fax:~(978)~689-3232.  E-mail: byl@psicorp.com, demers@psicorp.com, and rietman@psicorp.com, respectively.}
\thanks{This material is based upon work supported by US Army Space \& Missile Command under Contract Number W9113M-07-C-0204. Any opinions, findings and conclusions or recommendations expressed in this material are those of the authors and do not necessarily reflect the views of US Army Space \& Missile Command.}%
}


\maketitle
\thispagestyle{fancy}

\begin{abstract}
Rapid identification of object from radar cross section (RCS) signals is important for many space and military applications.  This identification is a problem in pattern recognition which either neural networks or support vector machines should prove to be high-speed.  Bayesian networks would also provide value but require significant preprocessing of the signals.  In this paper, we describe the use of a support vector machine for object identification from synthesized RCS data.  Our best results are from data fusion of X-band and S-band signals, where we obtained 99.4\%, 95.3\%, 100\% and 95.6\% correct identification for cylinders, frusta, spheres, and polygons, respectively.  We also compare our results with a Bayesian approach and show that the SVM is three orders of magnitude faster, as measured by the number of floating point operations.
\end{abstract}

\section{Introduction}
Classification of dynamic radar signals from exo-atmospheric objects is an important problem for military and space applications. Obviously different shapes and sizes can have significantly different meaning. Pebble-sized objects have little direct military interest as potential missiles.  However, pebble-sized objects can be of great interest for space craft survival. To keep our problem simple and to allow us to explore a new classification algorithm, we will focus on four types of objects; cylinders, frusta, spheres, and irregular flat polygons, ranging in size from centimeters to meters.

Geometric shapes of exo-atmospheric objects can be determined by analysis of the returned radar cross section (RCS) signatures~\cite{B:knott}.  Though some work has been done on computing the geometry from RCS~\cite{B:borden}, the approach is generally not effective.  Because the RCS signatures are unique patterns for different geometric types, this suggests using machine learning techniques for pattern recognition.  A vast literature exists for pattern recognition of many types of signals, signatures, and images.  Some of the most widely used are neural networks, Bayesian networks, and support vector machines (SVM). To use these pattern recognition systems generally requires a database of pairs of signatures and classes.  In our case this would be a database of synthesized RCS returns from representative frusta, cylinders, spheres, and polygons.

Machine learning techniques generally require two stages: one for training a system to learn how to separate geometric classes and a second for classifying new data as it arrives.  The data used for our studies are simulated RCS returns from multiple geometric shapes.  Accuracy is represented as an average of the percent correct, while precision is represented by measuring the standard deviation from a batch of runs consisting of training followed by testing and re-randomizing.

Depending on the desired application, we may require pattern recognition from RCS signatures be done quickly so that decisions for dealing with this information can be made in a timely manner.  This requires that the computational burden be minimized while maintaining the highest degree of accuracy.  The three common pattern recognition methods mentioned above could potentially be fast enough for these applications.  Each approach has its own advantages and disadvantages.

All three methods require a database of labeled signatures.  Generally, neural networks require at least three times as many data samples (preferably ten times as many) for training as the input dimensionality.  So if our RCS signatures consist of 1,000 data points, then we need 10,000 labeled signatures. This is potentially not a problem for synthesized databases.  However, as the input dimensionality increases, the number of samples goes up.  In many cases this is, or will quickly become, a problem.  During training of the neural network, there is invariably a good chance the learning algorithm will converge to a local minimum in the nonlinear hyperspace of the weights.  This can result in poor generalization or poor performance on testing and evaluation samples.  Neural networks could be very fast for RCS pattern recognition because once the network is trained, the evaluation simply consists of a few vector-matrix multiplications and some function applications to vectors.  Specifically, if we used a single hidden layer perceptron with 1,000 inputs and ten nodes in the hidden layer and one output, a new RCS classification would consist of multiplication of one 1000-element vector with a $1,000 \times 10$ element matrix to produce a ten-element vector to which we then apply a hyperbolic tangent for each element, followed by another vector-matrix product, in this case a ten-element vector by a $10 \times 1$ element matrix. The total would be on the order of 10~MFLOPS.  Clearly the neural network could provide good speed, but the performance is not adequate for most applications, especially applications requiring significant accuracy.  Neural networks also have the advantage that significant preprocessing or feature extraction of the RCS signatures may not necessarily be required. Farhat~\cite{B:farhat} conducted early research on using neural networks for automated target identification, but not from RCS.  His approach was more concerned with image reconstruction from microwave signatures and image recognition.  At the image recognition step, he applied the neural networks.  A good reference of neural networks and their use is~\cite{B:reed}.

An alternative pattern recognition method that could be used for our RCS analysis is a Bayesian network~\cite{B:berger}.  The Bayesian approach also requires a massive amount of data for training.  It also requires pre-computation of the moments of the distributions of the data.  This would necessarily require a, perhaps massive, reduction in dimensionality to apply Bayesian techniques to spectral identification and RCS signature classification.  The approach typically involves extracting some features from the signatures or spectra and using that population of features for the computation by the Bayesian methods.  As we will see, the pre-calculation of features often is quite time-consuming.

This paper reports our use of support vector machine (SVM)~\cite{B:cristianini}\cite{B:vapnik}.  Similar to the other methods, it requires a database of labeled RCS signatures.  The SVM has a number of distinct advantages over the other two techniques.  Unlike the neural network, it is not too likely that the training will get stuck in some nonlinear space, because the data are first transformed to a linear space, albeit, perhaps of very high-dimensionality.  Consequently there is a global minimum and the performance results can be far better than a neural network. Secondly, unlike the Bayesian network the SVM can accept very high-dimensional input vectors without preprocessing.

In the following sections we discuss construction of an RCS database for frusta, spheres, cylinders and polygons. We then discuss an SVM for RCS signature classification and compare the results with a Bayesian approach.

\section{\label{S:rcs}RCS Data Synthesis}

Three synthetic data sets were used to train and test the SVM.  The data sets ranged from simple radar cross section (RCS) versus viewing aspect angle tables to complex simulations of scenarios incorporating multiple radars operating at different frequencies and locations.  Most of the results presented in this paper are based upon a group of data sets of intermediate complexity, incorporating RCS versus viewing aspect for four object shapes of varying size and viewing aspect angle at two commonly used radar wavelengths, 3-GHz (S-band) and 10-GHz (X-band). The four shapes evaluated are:

\begin{enumerate}
\item Spheres with radii varying from 0.001--2 m.
\item Cylinders with diameters from 0.5--2 m and lengths from 1--12 times the diameter (1--20 m range).
\item Frusta, blunt nosed cones, with heights (the distance from the nose to the tail) of 0.5--2 m, tail diameters from 25--100\% the height, and nose diameters of 5--30\% the tail diameter.
\item Flat plate polygons with 3 to 5 edges with a maximum feature size of 0.3--6 m.
\end{enumerate}

\subsection{Radar Cross Section}

The RCS, $\sigma(\lambda, r)$, for each of these objects is a function of frequency, $\lambda$, and, for all objects except for the sphere, viewing aspect angle $\theta$.  The RCS models for the objects were drawn from a number of sources and each is unique. The RCS model for the sphere was adapted from~\cite{B:mahafza} and includes three regions:

\noindent Optical 
\begin{eqnarray}
\sigma =\pi r^{2}, & \; \; & r \gg \lambda 
\end{eqnarray}
\noindent Rayleigh 
\begin{equation}
\sigma \approx 9\pi r^{2}(kr)^{4}:\mbox{where }k=\frac{2\pi }{\lambda }
\mbox{ and }r\ll \lambda 
\end{equation}
and the Mie region, where the RCS is approximated using spherical Bessel functions.

\begin{figure}[htb]
\begin{center}
\scalebox{0.69}{\includegraphics{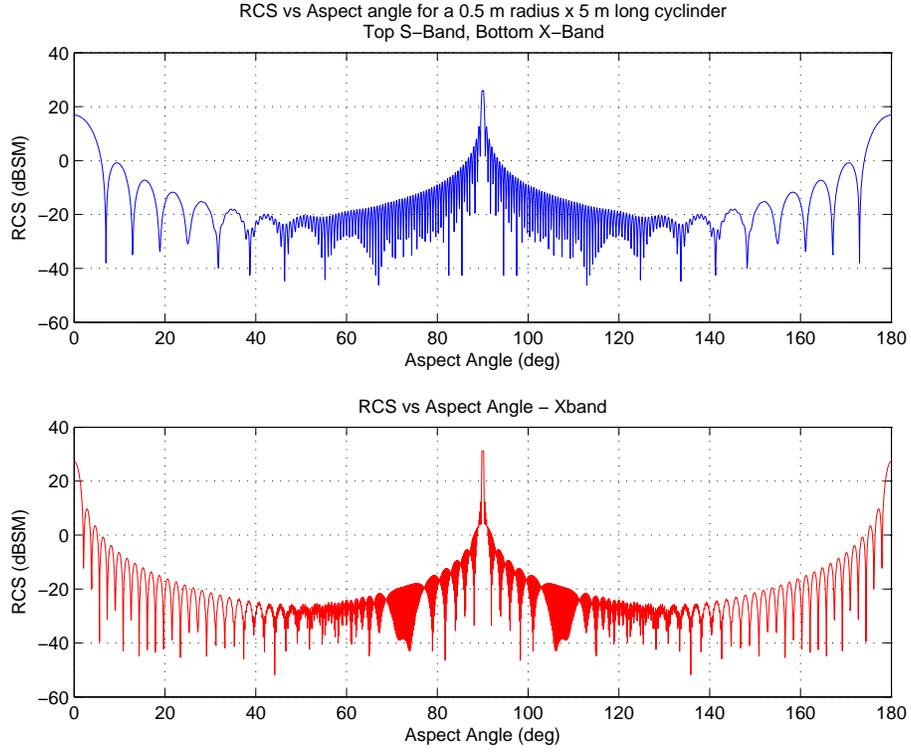}}
\caption{\label{F:cyl}RCS versus aspect angle for a 0.5~m by 5~m long cylinder
at both (a)~S-band and (b)~X-band.}
\end{center}
\end{figure}

The RCS model of a right cylinder used in this study is  
\begin{equation}
\sigma=kal^2\left|\cos\theta_i \frac{sin(kl\sin\theta_i)}{kl\sin\theta_i}
\right|^2  \label{eqn:cyl}
\end{equation}
where $a$ is the radius of the cylinder, $l$ is the cylinder length, and $\theta_i$ is the angle from broadside incidence. This solution is derived in Knott~\cite{B:knott} using physical optics. Because this solution becomes singular when $ \theta_i=0$, we substitute in the exact broadside solution
\begin{equation}
\sigma=kal^2
\end{equation}
when $\theta_i=90\pm 0.2^\circ$.  Because equation \ref{eqn:cyl} does not account for the cylinder end caps, we employ the theory of superposition and add the RCS of a circular plate
\begin{equation}
\sigma_{cap}=\pi k^2 a^4\left(\frac{2J_1(2ka\sin\theta)}{2ka\sin\theta}
\right)^2(\cos\theta)^2,\: \theta\neq 0
\end{equation}
and  
\begin{equation}
\sigma=\frac{4\pi^3a^4}{\lambda^2},\:\theta=0,
\end{equation}
where $ \theta=0 $ corresponds to normal incidence~\cite{B:mahafza}. Figure~\ref{F:cyl} shows both the S and X band RCS versus aspect angle for a 0.5~m by 5~m long cylinder.

\begin{figure}[htb]
\begin{center}
\scalebox{0.15}{\includegraphics{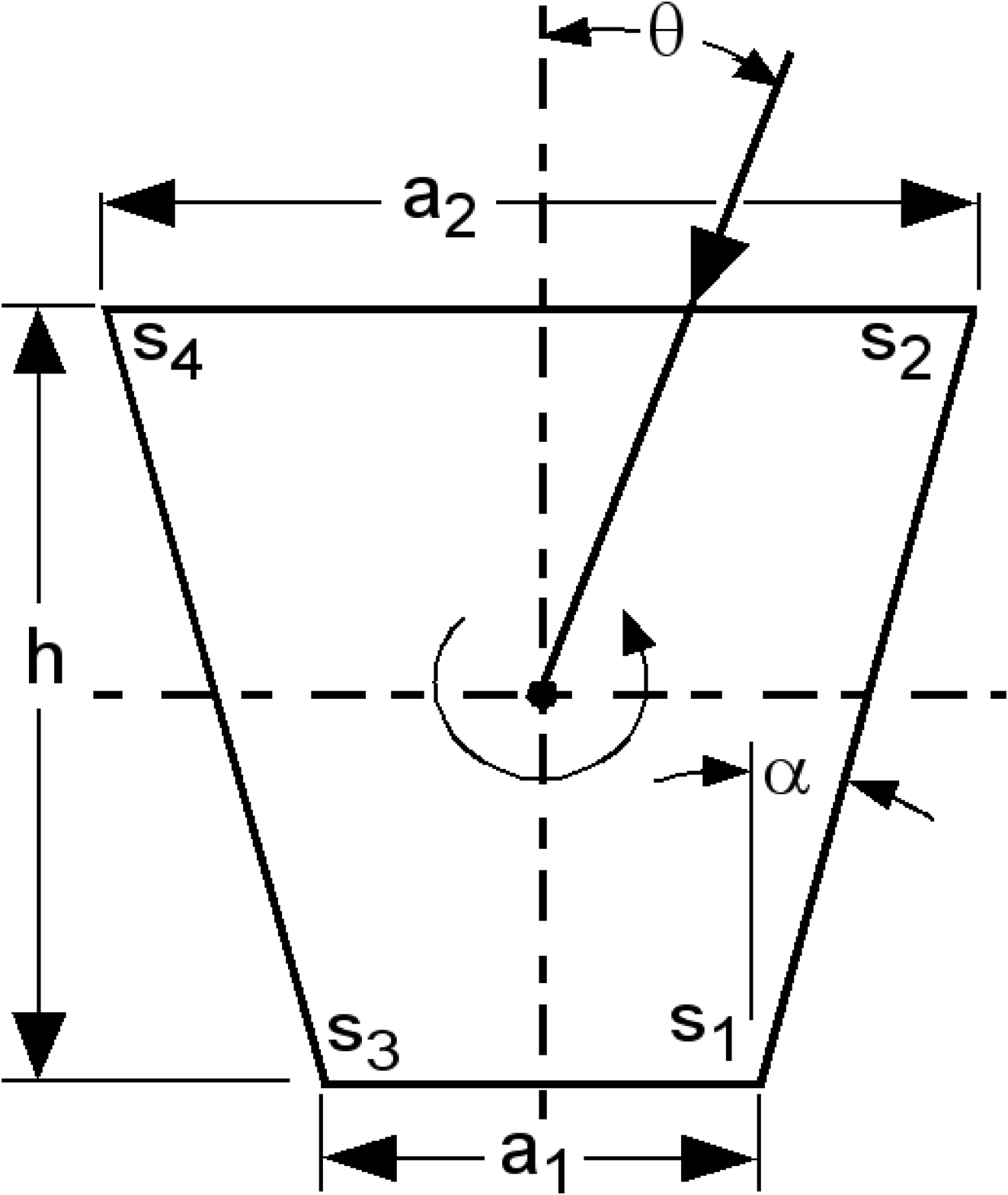}}
\caption{\label{F:scatter}Scattering centers geometry for frustum RCS.  The axis is of rotation is normal to the plane of the paper, centered at the origin of the dashed axes.}
\end{center}
\end{figure}

\begin{figure}[htb]
\begin{center}
\scalebox{0.69}{\includegraphics{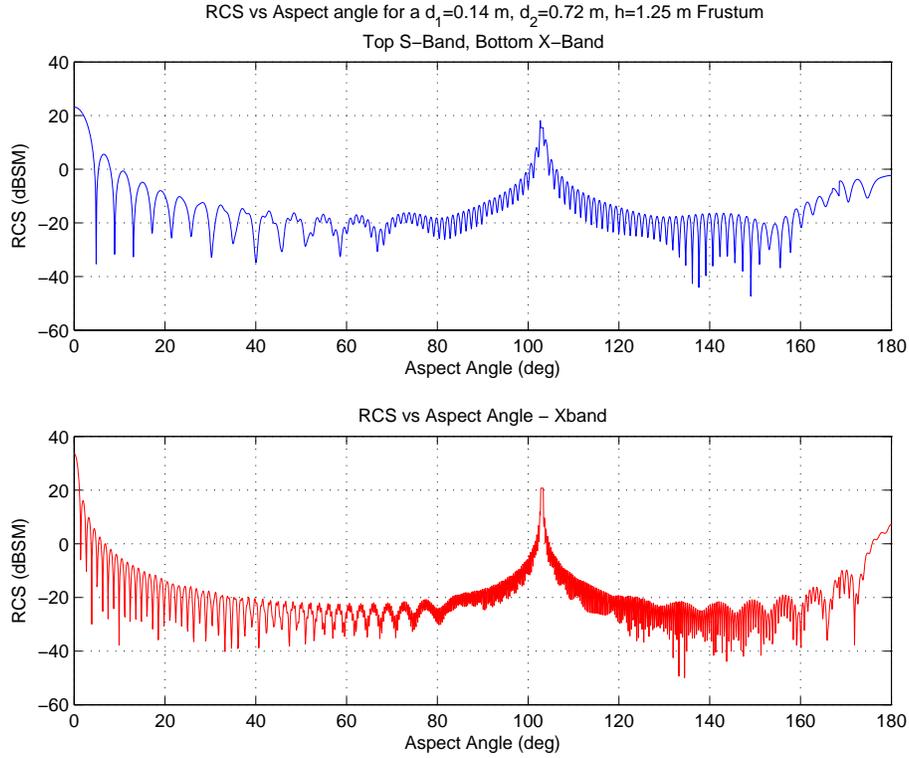}}
\caption{\label{F:frustum}RCS versus aspect angle for a frustum at both (a)~S- and (b)~X-bands.}
\end{center}
\end{figure}

For the frustum, we elected to use a classic set of formulas from~\cite{B:ross}, which utilizes the geometric theory of diffraction to predict the RCS of a frustum using 4 scattering centers:
\begin{equation}
\sqrt{\sigma}e^{j\rho}=\sum_{i=1}^4\sqrt{\sigma_i}e^{j\rho_i}.
\end{equation}
Figure~\ref{F:scatter} shows the locations of the four scattering centers and the key dimensions defining the frustum model.
Assuming that the radar is monostatic, the contribution of the 4 scatterers is
\begin{eqnarray}
\sqrt{\sigma_1}&=&\left\{
\begin{array}{cc}
g_1 \left[
\left(m_1-\cos\left(\frac{\pi+2\theta}{\eta_1}
\right)\right)^{-1}\mp\left(m_1-1 \right)^{-1}\right] & 
\theta\leq\pi-\alpha \\ 
0 & \theta>\pi-\alpha
\end{array}
\right.  \label{eqn:sigma1} \\
\sqrt{\sigma_2}&=&\left\{
\begin{array}{cc}
g_2 \left[
\left(m_2-\cos\left(\frac{3\pi-2\theta}{\eta_2}
\right)\right)^{-1}\mp\left(m_2-1 \right)^{-1}\right] & 
\theta\geq-\alpha \\ 
0 & \theta<\-\alpha
\end{array}
\right.  \label{eqn:sigma2} \\
\sqrt{\sigma_3}&=&\left\{
\begin{array}{cc}
g_1 \left[
\left(m_1-\cos\left(\frac{\pi-2\theta}{\eta_1}
\right)\right)^{-1}\mp\left(m_1-1 \right)^{-1}\right] & 
\theta\leq\pi/2 \\ 
0 & \theta>\pi/2
\end{array}
\right.  \label{eqn:sigma3} \\
\sqrt{\sigma_4}&=&\left\{
\begin{array}{cc}
g_2 \left[
\left(m_2-\cos\left(\frac{3\pi+2\theta}{\eta_2}
\right)\right)^{-1}\mp\left(m_2-1 \right)^{-1}\right] & 
\theta\leq-\alpha \\ 
0 & \pi/2>\theta>\alpha \\ 
g_2 \left[
\left(m_2-\cos\left(\frac{-\pi+2\theta}{\eta_2}
\right)\right)^{-1}\mp\left(m_2-1 \right)^{-1}\right] & 
\theta\leq\pi/2
\end{array}
\right.  \label{eqn:sigma4}
\end{eqnarray}
with
\begin{eqnarray}
g_1 & = & \frac{\sin(\pi/\eta_1)}{\eta_1}\sqrt{\frac{a_1\csc\theta}{k}} \\
g_2 & = & \frac{\sin(\pi/\eta_2)}{\eta_2}\sqrt{\frac{a_2\csc\theta}{k}} \\
m_1 & = & \cos(\pi/\eta_1) \\
m_2 & = & \cos(\pi/\eta_2)
\end{eqnarray}
and the phases referenced to the base of the frustum are given by 
\begin{eqnarray}
\rho_1&=&-2k[a_1\sin\theta+2h\cos\theta]+\pi/2 \\
\rho_2&=&-2ka_2\sin\theta+\pi/4 \\
\rho_3&=&2k[a_1\sin\theta-2\cos\theta]-\pi/4 \\
\rho_4&=&2ka_2\sin\theta-\pi/4
\end{eqnarray}
where $\alpha$ is the frustum angle in radians 
\begin{equation}
\alpha=\tan^{-1}\left(\frac{a_2-a_1}{2h}\right)
\end{equation}
and 
\begin{eqnarray}
\eta_1&=&\frac{3}{2}-\frac{\alpha}{\pi} \\
\eta_2&=&\frac{3}{2}+\frac{\alpha}{\pi}
\end{eqnarray}
and where $\theta$ is the viewing aspect as measure along the axis of symmetry from the large end of the frustum ($a_2$). The choice of signs in equations [\ref{eqn:sigma1}]-[\ref{eqn:sigma4}] relate to the polarization convention (upper sign for vertical polarization, lower sign for horizontal polarization). For this analysis, we assumed vertical polarization. As is the case for the cylinder, this solution has singularities at $\theta=0,\pi/2-\alpha,\pi$. These singularities have been handled as described in~\cite{B:ross}.  Figure~\ref{F:frustum} shows both the S- and X-band RCS versus aspect angle for a frustum where $d_1=0.14$~m, $d_2=0.72$~m, and $h=1.25$~m, as shown in Figure~\ref{F:frustum}.

\begin{figure}[htb]
\begin{center}
\scalebox{1.0}{\includegraphics{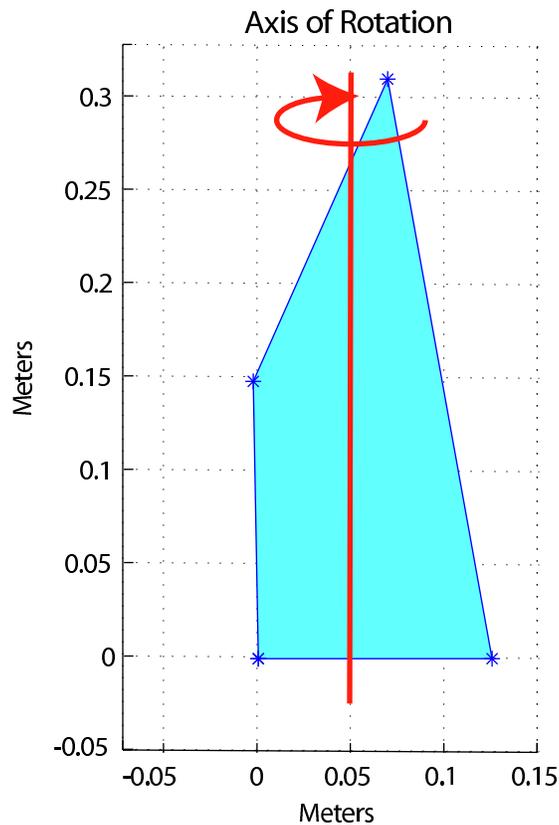}}
\caption{\label{F:poly}A typical five sided polygon.}
\end{center}
\end{figure}

\begin{figure}[htb]
\begin{center}
\scalebox{0.69}{\includegraphics{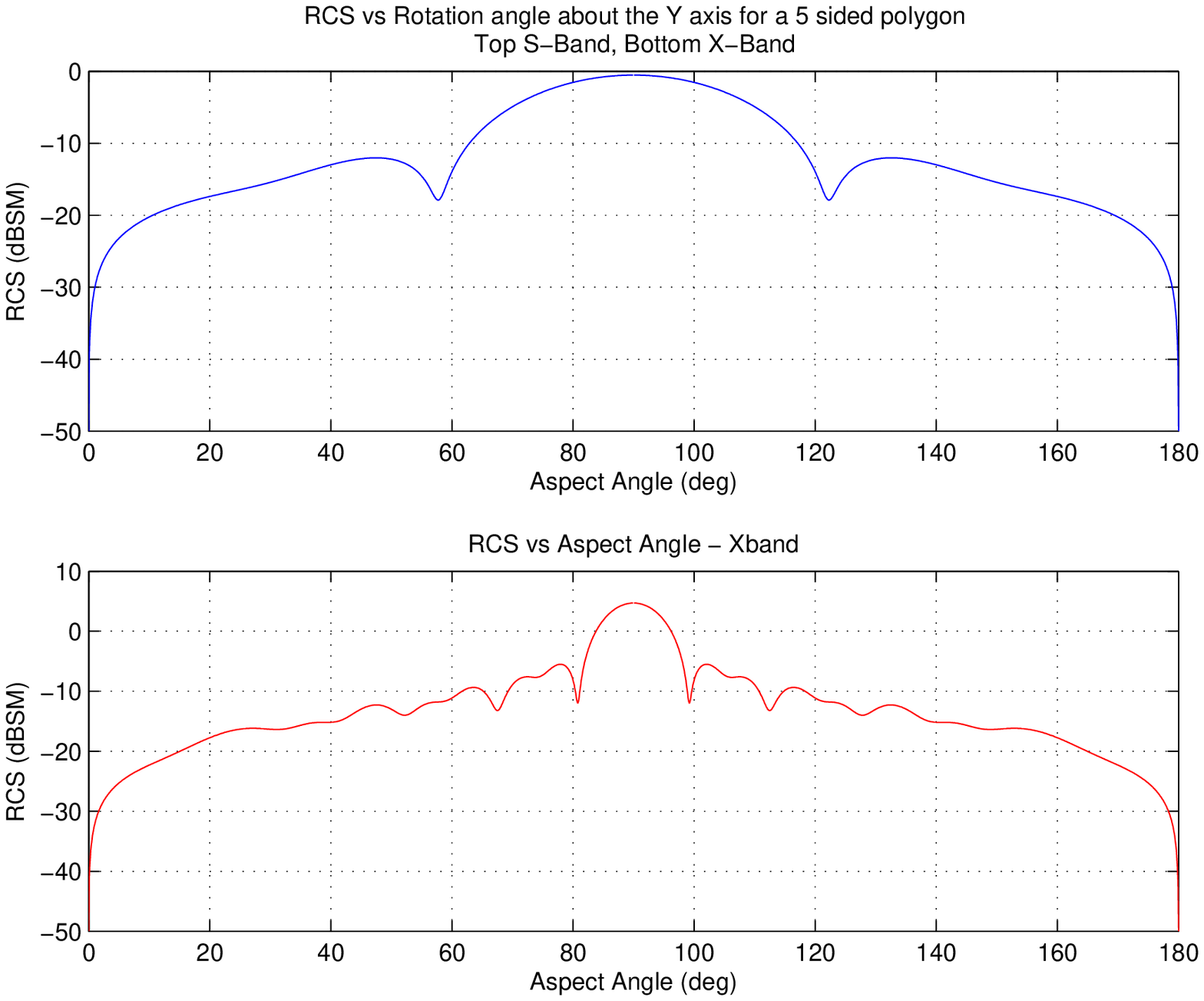}}
\caption{\label{F:polyrcs}RCS versus aspect angle for the polygon shown in Figure~\protect\ref{F:poly}, for (a)~S-band and (b)~X-band.  Axis of rotation shown in Figure~\protect\ref{F:poly}.}
\end{center}
\end{figure}

Lastly for the $N$-sided polygon, we applied a physical optics solution from~\cite{B:mahafza}, where the expression for an arbitrary $N$-sided polygon in a local co\"{o}rdinate frame has been evaluated analytically as
\begin{equation}
S(u,v)=\sum^N_{n=1}e^{j(\boldsymbol{\omega}\cdot\boldsymbol{\gamma})}\left[\frac{\hat{n}
\times\hat{\alpha}_n\cdot\hat{\alpha}_{n-1}} {(\boldsymbol{\omega}\cdot\boldsymbol{\alpha}
_n)(\boldsymbol{\omega}\cdot\boldsymbol{\alpha}_{n-1})}\right]
\end{equation}
where $\boldsymbol{\gamma}_n$ are the polygon vertices and $\alpha_n$ are the edge
vectors given by 
\begin{equation}
\boldsymbol{\alpha}_n=\frac{\boldsymbol{\gamma}_{n+1}-\boldsymbol{\gamma}_n}{|\boldsymbol{\gamma}_{n+1}-
\boldsymbol{\gamma}_n|}
\end{equation}
and $\boldsymbol{\omega}=u\hat{x}+v\hat{y}$. Figure~\ref{F:poly} shows a randomly
generated five-sided polygon, typical of those generated for use in this
study. Figure~\ref{F:polyrcs} shows the calculated RCS for the polygon
shown in Figure~\ref{F:poly} when it is rotated about the axis shown in
Figure~\ref{F:poly} ($\theta=0,180^\circ$ represent the edge on condition).


\subsection{Training Sets}

Now that we have defined the objects that are going to be incorporated into data sets we need to use these models to generate simulated radar data.  Ideally we would like to train the SVM using data similar to what would be observed in an actual engagement but for simplicity we started with a scenario similar to what one would encounter on a radar test range. The first set of data consisted of 10,000 randomly generated frusta and cylinders whose RCS versus aspect angle were calculated using 2,000 evenly spaced samples from 0--180$^\circ$. 

\begin{figure}[htb]
\begin{center}
\scalebox{0.68}{\includegraphics{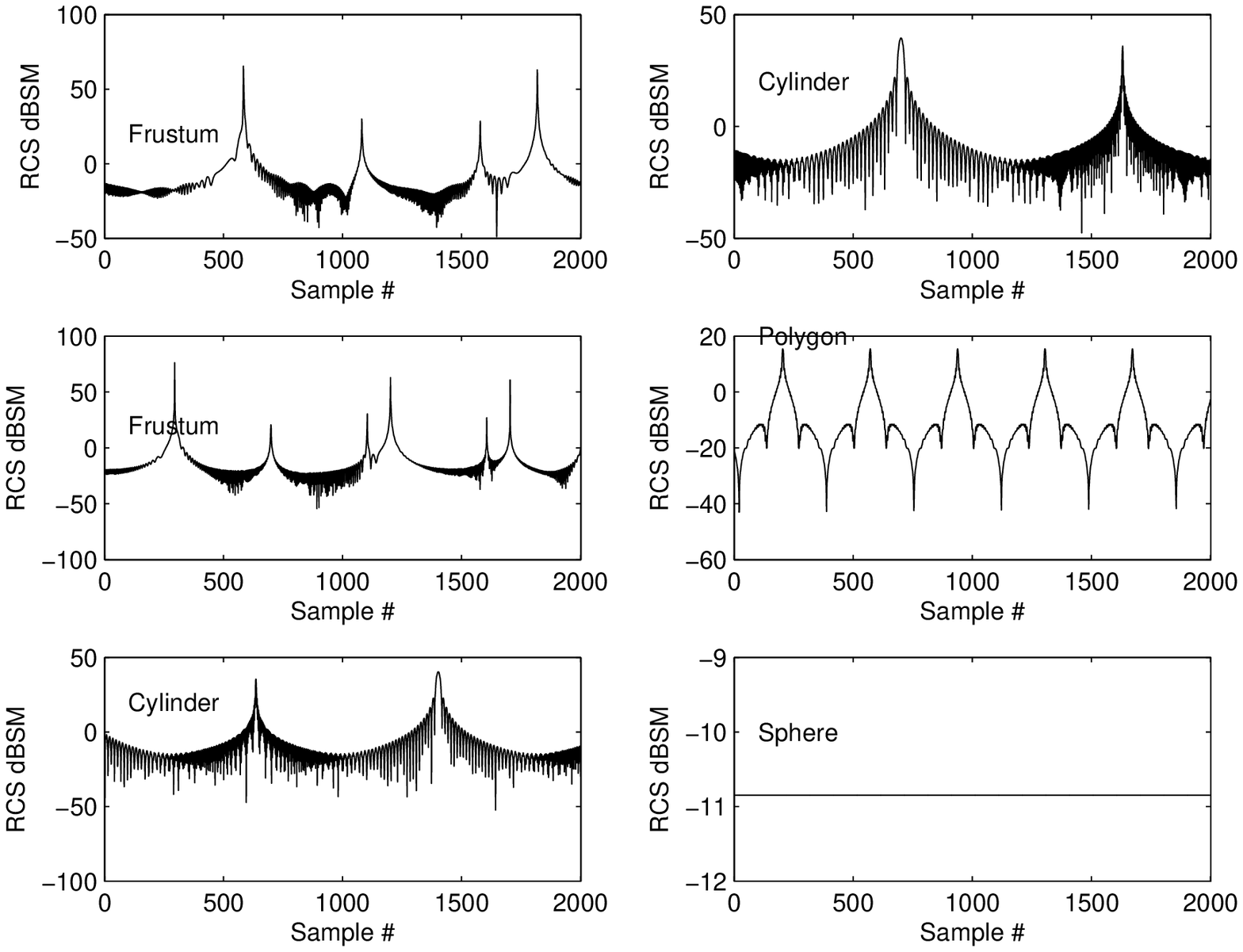}}
\caption{\label{F:train}RCS versus sample (aspect angle) for six objects in the final data set.  From upper-left to lower-right, moving left-to-right: (a)~frustum, (b)~cylinder, (c)~frustum, (d)~polygon, (e)~cylinder, and (f)~sphere.}
\end{center}
\end{figure}

Once the performance of the SVM had been verified using this data set, we increased the difficulty of the problem by varying the aspect through which each object was rotated. The second data set consisted of randomly generated frusta, cylinders, spheres, and polygons rotated through an aspect starting at 0$^\circ$ and ending at $180^\circ$ (the RCS was calculated at 2,000 evenly distributed points through this rotation).  There were 5,000 data vectors for each object type, resulting in a total of 20,000 data vectors.  The final data set consisted of 20,000 randomly generated frusta, cylinders, spheres, and polygons rotated from a random start angle with a minimum rotation of 180$^\circ$ and a maximum rotation of 1080$^\circ$.  Figure~\ref{F:train} shows the RCS versus sample for six objects in the final data set, with sample being the sampled aspect angle.

Summarizing, our data consisted of three data sets.  The first data set comprised 4930 cylinders and 5070 frusta for a total of 10,000 objects, approximately a 50\% distribution between the two classes. The data for each object consisted of 2,000 RCS values, starting at broad-side specular to an observing monostatic radar and rotating one-half revolution. The axis of rotation was normal both to the axis of symmetry of the object and the direction of observer. The frequency was selected at random from one of the following four frequencies: 900~MHz, 3~GHz, 6~GHz, and 10~GHz, simulating four different viewers. The RCS can be obtained from a radar return given an object's range and the gain from the radar station, and varying it across azimuth simulates the changing perspective an observing ground radar would have of dynamic objects.

The second data set comprised objects from four different classes: cylinders, frusta, spheres, and random polygonal plates. The data for each object consisted of 2,000 radar cross-section (RCS) values for each object at a $0^\circ$-placement relative to an observing mono-static radar and rotating at a random rate with revolution over the 2,000 values ranging between one and five periods. Five thousand objects were generated in each of the four classes, with varying lengths and radii, resulting in 20,000 total objects.  Again the data set consisted of the four frequencies for data set~1.

The third data set replicated the second data set, except that the starting angle to each observer was chosen randomly.  Also, the returns for each object at S-band (3~GHz) and X-band (10~GHz) frequencies, effectively simulated the returns from two separate ground radars observing the same object.  This effectively doubles the amount of data for each object, \emph{i.e.,} the length of each vector. Our goal was to combine the two radar signatures, via sensor fusion, to improve classification and identification.

\section{Background to Support Vector Machines}

Having synthesized the RCS data, we are ready to begin the processing with the support vector machine (SVM).  The following theoretical outline is similar to Vapnik~\cite{B:vapnik}.  Given our set of labeled training pairs of RCS signatures and class labels for what they represent we can write this as follows: $(\boldsymbol{x_{1}},y_{1}),$ $(\boldsymbol{x_{2}},y_{2}),$ $(\boldsymbol{x_{3}},y_{3}),$ $\ldots $, where $\boldsymbol{x_{i}}$ represents the RCS signature and $y_{i}$ represents the class label (sphere, frustum, cylinder, polygon).  To simplify the problem, we will label an RCS signature as $+1$ if it represents a cylinder and $-1$ if it represents anything else.  We thus partition our data set into four labeled groups, depending on the object type.  As described above in Section~\ref{S:rcs}, each object-type data set consists of 5,000 RCS signatures. We can now form a separating hyperplane as follows:

\begin{equation}
\begin{array}{ccc}
w-x_{i}+b\geq 1 & \text{if} & y_{i}=1 \\ 
w-x_{i}+b<1     & \text{if} & y_{i}=-1%
\end{array}
\end{equation}
where we see the output classes are $-1$ and $+1$. The inequalities in the above equation can be written as:
\begin{equation}
\begin{array}{cc}
y_{i}(w\cdot x_{i}+b)\geq 1, & i=1,\ldots, n
\end{array}
\end{equation}
for $n$ training samples. The relations in this equation become constraints for the optimal hyperplane given by
\begin{equation}
w_{0}\cdot x+b=0
\end{equation}
where $b$ is a constant known as the bias.  This is the unique hyperplane for maximum separation of the training data.  The vectors (RCS signatures) for which the left-hand-side is equal to 1 are called the support vectors. The optimal separating hyperplane is:
\begin{equation}
w_{0}=\sum_{i=1}^{n}y_{i}\alpha _{i}^{o}x_{i}
\end{equation}%
where $\alpha _{i}^{o}\geq 0$. If we let $\Lambda _{0}^{T}=(\alpha _{1}^{0},\ldots ,\alpha _{n}^{0})$ then we could solve the quadratic programming problem. Basically we want to maximize
\begin{equation}
w(\Lambda ,\delta )=\Lambda ^{T}\underline{1}-\frac{1}{2}\left[ \Lambda
^{T}D\Lambda +\frac{\alpha _{\max }^{2}}{c}\right] 
\end{equation}
subject to the constraints
\begin{eqnarray}{c}
\Lambda ^{T}Y & = & 0 \\ 
\delta & \geq & 0 \\ 
0 \leq & \Lambda & \leq \delta \boldsymbol{u}
\end{eqnarray}
where $Y^{T}=(y_{1},\ldots ,y_{n})$ is a vector of labels; $\boldsymbol{u^{T}} $ is a unit vector; $D$ is an $n \times n$ matrix with elements given by $D_{ij}=y_{i}y_{j}x_{i}x_{j}$ with $i,j=(1,\ldots ,n)$; and $\alpha _{\max }=\max (\alpha _{1},\ldots ,\alpha _{n})$ are the weights for the support vectors and $c$ is a constant analogous to bias. The classifier function is now
\begin{eqnarray}
f(x) & = & w\cdot \phi (x)+b \\
w    & = & \sum\limits_{i=1}^{n}y_{i}\alpha _{i}\phi (x_{i})
\end{eqnarray}
The values $\phi (x_{i})$ are the transformed objects in feature space. We can rewrite the function as
\begin{eqnarray}
f(x) & = & \phi (x_{i})\cdot w+b  \nonumber \\ 
     & = & \sum\limits_{i=1}^{n}y_{i}\alpha _{i}\phi \left(x_{i}\right) \cdot \phi (x_{i})+b
\end{eqnarray}
or more compactly as
\begin{eqnarray}
f(x) & = & \sum\limits_{i=1}^{n}y_{i}\alpha _{i}K(x,x_{i})
\end{eqnarray}

This kernel representation is the conventional way of representing a support vector machine.  As pointed out above, we could find the support vector weights, $\alpha _{i}$, by quadratic programming.  Instead, we will use the perceptron algorithm when combined with an SVM, also known as a kernel adatron. The algorithm is basically a gradient descent in error space to find the support vectors.  Veropoulus~\cite{B:veropoulos} is a good reference for the kernel adatron method.

There are several available methods for computing kernels for SVMs. These include the linear kernel, the polynomial kernel, the radial basis function (RBF) kernel, and the hyperbolic tangent kernel.  In this analysis, only the linear kernel was used.  The linear kernel is computed on the $L \times N$ matrix containing the training vectors as a covariance matrix calculation, $\widehat{K}=LL^{T}$ where $K$ is a symmetric, positive definite matrix. The resulting kernel can then be normalized by:
\begin{eqnarray}
D & = & \sqrt{diag(K)} \\
K & = & D\widehat{K}D
\end{eqnarray}

Once $K$ has been computed, a gradient descent process is used to determine whether or not each vector is a support vector.  $S$ support vectors are selected from the initial $L$ training vectors. After the learning process converges the location of the non-zero elements in the $\boldsymbol{\alpha}$ vector, an $L$-dimensional vector, are pointers to the support vectors. The actual value of the non-zero elements in this $\boldsymbol{\alpha}$ vector are the coefficients for the testing phase.

A vector under test, in this case an RCS signature, $\boldsymbol{v}$, can be classified by computing the dot-product of $\boldsymbol{v}$ with each support vector, $\boldsymbol{s}$, scaling each result by $\alpha$ and the response, $y$, of each support vector. The responses for the support vector are $+1$ for positive cases and $-1$ for negative cases. The vector under test is evaluated according to:
\begin{eqnarray}
z & = & \sum\limits_{j=1}^{S}\alpha _{j}y_{j}\boldsymbol{v}\cdot\boldsymbol{s}
\end{eqnarray}
If $z>0$, the object under test is classified as belonging in the set; otherwise it is classified as being outside the set.

\section{Training and Testing}

From the 10,000 objects in the first data set, two trials of training took place: one with 400 randomly selected objects, with testing on the remaining 9,600; and one with 4,000 randomly selected objects, with testing on the remaining 6,000. All tests were repeated ten times (ten-fold cross-validation) to obtain statistical measurement for accuracy and precision.

From the 20,000 objects in the second set, 4,000 were randomly selected for training the kernel adatron. The remaining 16,000 objects were then tested against the support vectors and the weight vector, $\boldsymbol{\alpha}$.  Tests were performed for each of these four classes using a separately-trained SVM classifier:

\begin{itemize}
\item Cylinder \emph{vs.} non-cylinder
\item Frustum \emph{vs.} non-frustum
\item Sphere \emph{vs.} non-sphere
\item Polygons \emph{vs.} non-polygons
\end{itemize}

\noindent All tests were repeated ten times.

From the 20,000 objects in the third data set, 8,000 were randomly selected for training the kernel. The remaining 12,000 objects were then tested against that kernel. Four tests were performed for each of the classes, as in the second data set. All the tests were repeated twenty times (twenty-fold cross-validation).

We used the kernel adatron algorithm presented on pages~80--81 in~\cite{B:veropoulos}, with fixed values of $\eta =0.01$, $\tau =500$, and a threshold of 5,000. These values affect the selection of the support vectors, and thus the classification results.  However, they remained fixed for the purpose of these numerical trials.  Table~\ref{T:desc} describes each experiment performed, including the data set used, the number of vectors used for training and testing, the length these each vectors compared to the ones in that data set, and other descriptive information.

\begin{table}[htb]
\begin{center}
\begin{tabular}{|rrrrlp{1.5in}|}
\hline
\textbf{exp.}   & \textbf{data} & \textbf{vectors} & \textbf{vectors} & \textbf{vector} & \textbf{brief} \\
\textbf{no.} & \textbf{set}  & \textbf{trained} & \textbf{tested}  & \textbf{length} & \textbf{description} \\
\hline \hline
1  & 1 &  400 &  9600 & full    & \\
2  & 1 & 4000 &  6000 & full    & \\
3  & 1 &  400 &  9600 & quarter & \\
4  & 1 & 4000 &  6000 & quarter & \\
5  & 1 & 4000 &  6000 & quarter & FFT \\
6  & 1 & 4000 &  6000 & quarter & FFT-w \\ \hline
7  & 2 & 4000 & 16000 & full    & FFTdc, 1--10~rot., random \\
8  & 2 & 4000 & 16000 & quarter & FFTdc, first quarter \\
9  & 2 & 4000 & 16000 & quarter & FFTdc, second quarter \\
10 & 2 & 4000 & 16000 & quarter & FFTdc, third quarter \\
11 & 2 & 4000 & 16000 & quarter & FFTdc, fourth quarter \\ \hline
12 & 3 & 8000 & 12000 & full    & FFTdc, X-band, S-band \\
13 & 3 & 8000 & 12000 & quarter & FFTdc, X-band, S-band, random \\
14 & 3 & 8000 & 12000 & quarter & FFTdc, X-band, S-band, random, moments\\
\hline
\end{tabular}
\caption{\label{T:desc}Summary of the experiments (see text for more details).}
\end{center}
\end{table}

\subsection{First Data Set}

We tested only for cylinders, so the non-cylinder class is equivalent to the frustum class because there are only two classes in this data set. Testing was performed in a variety of ways. First, the raw RCS over the varying azimuths was used, with the full 2,000 samples per object.  Training on 400 random objects and testing on 9600 objects resulted in $99.4\%\pm 0.9\%$, decomposed as shown in experiment~1 of Table~\ref{T:results1}.  Training on 4,000 random objects and testing on 6,000 random objects resulted in $100.0\%\pm 0.0\%$, decomposed as shown in experiment~2 of Table~\ref{T:results1}.

\begin{table}[htb]
\begin{center}
\begin{tabular}{|r|r@{ }r|r@{ }r|r@{ }r|r@{ }r|r@{ }r|r@{ }r|r@{ }r|}
\hline
\textbf{exp.}   & \multicolumn{2}{c|}{\textbf{cylinder}} & \multicolumn{2}{c|}{\textbf{frustum}} & \multicolumn{2}{c|}{\textbf{total correct}} \\
\textbf{number} & \textbf{$\mu$} & \textbf{$\sigma$} & \textbf{$\mu$} & \textbf{$\sigma$} & \textbf{$\mu$} & \textbf{$\sigma$} \\
\hline \hline
1  & 49.1 & (1.4) & 50.3 & (1.3) & 99.4 & (0.9)  \\ 
2  & 49.3 & (0.0) & 50.7 & (0.0) &100.0 & (0.0)  \\ 
3  & 45.9 &(12.2) & 49.7 & (3.1) & 95.6 & (5.7)  \\ 
4  & 47.6 & (3.8) & 50.2 & (4.3) & 97.8 & (2.5)  \\ 
5  & 45.7 & (6.4) & 50.1 & (0.6) & 95.9 & (2.9)  \\ 
6  & 47.1 & (4.2) & 48.7 & (8.4) & 95.6 & (3.5)  \\ \hline
\end{tabular}
\caption{\label{T:results1}The result of the experiments, where $\mu$ is the mean correct and $\sigma$ is the standard deviation when run multiple times (see text for more details).  All units are percentages.}
\end{center}
\end{table}

\begin{table}[htb]
\begin{center}
\begin{tabular}{|r|r@{ }r|r@{ }r|r@{ }r|r@{ }r|r@{ }r|r@{ }r|r@{ }r|r@{ }r|}
\hline
\textbf{exp.}   & \multicolumn{2}{c|}{\textbf{cylinder}} & \multicolumn{2}{c|}{\textbf{frustum}} & \multicolumn{2}{c|}{\textbf{sphere}} & \multicolumn{2}{c|}{\textbf{polygon}} \\
\textbf{number} & \textbf{$\mu$} & \textbf{$\sigma$} & \textbf{$\mu$} & \textbf{$\sigma$} & \textbf{$\mu$} & \textbf{$\sigma$} & \textbf{$\mu$} & \textbf{$\sigma$} \\
\hline \hline
7  & 99.6 & (0.3) & 93.2 & (7.7) & 100  & (0.0)  & 83.8 & (14.2) \\ 
8  & 76.8 & (5.2) & 74.8 & (7.7) & 100  & (0.0)  & 76.5 & (22.4) \\ 
9  & 93.8 & (2.9) & 73.0 & (9.7) & 95.5 & (9.5)  & 79.5 & (16.6) \\ 
10 & 91.4 & (3.4) & 84.3 & (7.1) & 100  & (0.0)  & 85.0 & (4.9)  \\ 
11 & 91.9 & (3.1) & 87.7 & (3.0) & 100  & (0.0)  & 72.1 & (16.1) \\ \hline
12 & 99.6 & (0.4) & 84.9 & (2.6) & 100  & (0.0)  & 88.3 & (2.4)  \\ 
13 & 91.8 & (2.1) & 78.2 & (9.6) & 100  & (0.0)  & 85.8 & (3.1)  \\ 
14 & 99.4 & (0.8) & 95.3 & (3.3) & 100  & (0.0)  & 95.6 & (4.1)  \\
\hline
\end{tabular}
\caption{\label{T:results2} The result of the experiments, where $\mu$ is the mean correct and $\sigma$ is the standard deviation when run multiple times (see text for more details).  All units are percentages, and reflect objects correctly identified in independent tests.}
\end{center}
\end{table}

When using just one-quarter of the 2,000 samples per object, corresponding to $45^{\circ }$ of revolution, results could be little better than a random guess (50\%) depending on the swath visible to the observer; these results are not included in Table~\ref{T:results1}.  At this point, testing began on data processed by a Fourier transform instead of the raw azimuthal RCS values.  Only the magnitude values of the data in the were used.  The data were normalized by the magnitude of the DC component (the first value).  Repeating the above tests resulted in $95.6\%\pm 5.7\%$ with 400 randomly selected training objects and 9600 testing objects, decomposed as shown in experiment~3 of Table~\ref{T:results1}.   Training over 4,000 objects and testing on 6,000 objects resulted in $97.8\%\pm 2.5\%$, with the results shown in experiment~4 of Table~\ref{T:results1}.

These results compare well to using the raw RCS values. When only a quarter of the RCS data were selected, results match those of the raw RCS, which depend on the visible swath of data to contain distinguishing characteristics.  However, an SVM is dependent on data being placed in the same place in the data vector. Because we will not know the aspect angle to the observer, raw RCS returns are not as useful.  Fourier-transformed data places information in the same feature bins of each data vector, and this now removes this issue for SVM training.

When a Kaiser window was applied to the data, the results do not change significantly.  Using 4,000 randomly-selected training samples, following a Fourier transform, of objects and testing on the remaining 6,000 objects, the results without the window were $95.9\%\pm 2.9\%$, as shown in experiment~5 of Table~\ref{T:results1}; while with the window, the results were $95.6\%\pm 3.5\%$, decomposed as shown in experiment~6 of Table~\ref{T:results1} (marked as FFT-w in Table~\ref{T:desc}).  This shows that windowing did not significantly affect the results.

\subsection{Second Data Set}

Testing with this set now shows results when the amount of rotation is not fixed to one revolution but instead corresponds to a random number of full revolutions (marked as 1--10 rot. in Table~\ref{T:desc}), up to ten, for each object. Now that we have four objects with equal populations in the data set, a random guess can be considered 75\% accurate because stating everything is not in the class would be correct 3/4 of the time. All testing is performed on data following a Fourier transform, with the DC component (marked as FFTdc in Table~\ref{T:desc}) normalized to unity.

Using the full-length 2,000-point RCS data, training on a randomly-selected 4,000 objects, and testing on the remaining 6,000 objects, the results were those shown in experiments~7 of Table~\ref{T:results2}.  The SVM correctly identified the objects as cylinders 99.6\% of the time, frusta 93.2\% of the time, spheres 100\% of the time, and polygons 83.8\% of the time.  Tests using the second and third data sets used the four separately-trained SVM classifiers.

Training on consecutive, one-quarter-length swaths and testing with one-quarter-length data swaths, gave the results shown in experiments~8--11 of Table~\ref{T:results2}.  The results from these tests make sense.  Keeping in mind that the starting point for all objects in this data set is at $0^{\circ }$, which is nose-on to the observer while $90^{\circ }$ is broadside, and the objects each rotate at random speeds, the first quarter contains little useful information to distinguish a cylinder from a frustum.  Compare the RCS returns of Figure~\ref{F:cyl} and Figure~\ref{F:frustum} for aspect angles $0^\circ$ -- $45^\circ$ to see the difficulty in correctly classifying objects using only the first quarter of the data.  Spheres are always easy to distinguish given the uniform RCS returns at every aspect angle.  Polygons are expected to be difficult to distinguish given they exhibit different RCS return behavior with their varying sizes and number of sides. Proceeding from quarter to quarter, the chance of including useful RCS information increases, albeit not linearly. The third data set corrects this by starting at a random aspect angle to the observer.

\subsection{Third Data Set}

All the data in the third data set was processed by a Fourier transform and normalized to unity DC.  Each object has two sets concatenated together, one for S-band returns and one for X-band returns. The third data set trains on 8,000 randomly selected objects. Using 2,000 RCS data points from both ground observers resulted in what is shown in experiment~12 of Table~\ref{T:results2}.  The objects were identified correctly as cylinders 99.6\% of the time, frusta 84.9\% of the time, spheres 100\% of the time, and polygons 88.3\% of the time.

When using a random swath of 500 consecutive time samples for each object, instead of the full 2,000, we get the results shown in experiment~13 of Table~\ref{T:results2}.  The results for the correct identification of objects fell slightly, with cylinders at 91.8\%, frusta at 78.2\%, spheres at 100\%, and polygons at 85.3\%.

The first four moments, mean, variance, skew, and kurtosis, were appended as data points to the Fourier-transformed vector for each object.  These are the first four moments of the RCS data normalized so the largest moment has a value of one (marked as moments in Table~\ref{T:desc}). These were computed for both the X-band and the S-band data sets, resulting in eight additional data points.  The results of repeating these tests with the augmented data for the quarter-length data are as shown in experiment~14 of Table~\ref{T:results2}.  These resulted in significantly improved identification, with cylinders at 99.4\%, frusta at 95.3\%, spheres at 100\%, and polygons at 95.6\%.

\section{Discussion and Conclusions}

We have shown that it is possible to obtain very good classification of synthesized RCS signatures following a Fourier transform, with little or no other preprocessing.  This reduction in computational load can have significant impact in some applications. This direct testing on Fourier-transformed data obviates handling the starting aspect angle of data collection; the additional features, such as the first four statistical moments of the RCS data (mean, variance, skew, and kurtosis), may be added to an input vector to improve accuracy. Table~\ref{T:results2}  contains a summary of the results for all the numerical experiments.  Accuracy is measured as an overall percentage correct, with ten-fold (data set~1) or twenty-fold cross-validation (data sets~2 and~3). In an actual scenario, knowledge of the aspect angle of the object to the observer would not be known. Also, there is no guarantee that a full revolution of the object will be viewed. Therefore, the results from experiments~13 and 14 are likely closest to actual. Because the overall results from experiment~14 are better, using the moments should provide enhanced performance.  Experiment~14 provides a start at reasonable results for an actual scenario.

In order to relate our SVM results with another technique, we compared them to a Bayesian network.  Bayesian networks are a popular technique for pattern recognition and classification~\cite{B:berger}\cite{B:frey}.  The basic concept is to combine conditional probabilities using Bayes' theorem.

Donald Maurer of Johns Hopkins University, Applied Physics Laboratory, has applied the Bayesian technique to the identification of cones and cylinders~\cite{B:maurer}.  His initial starting point was RCS data from which he extracted several features.

As described in Maurer's paper, he used data of this type to build the Bayesian network. Data were processed as ten-degree aspect angle increments. The best overall performance by the Bayesian approach classifies cylinders from cones correctly 99\% of the time~\cite{B:maurer}.

For comparison, we used Maurer's data and processed it by a SVM. The features of the data were analyzed using the kernel adatron.  The four features that most contributed to identification in a Bayesian classifier were selected from both the cone and cylinder data sets~\cite{B:maurer}.  These comprise one object each, with features from azimuths in the range 0 to 180$^\circ$ in one-tenth-degree increments. Ten-degree swaths of each of the four features were used to compose the data for each vector, giving each vector a length of 400 features.  1700 vectors were created for the cone, and 1700 for the cylinder.  The kernel adatron trained on 1700 randomly selected vectors, and tested on the remaining 1700, in a twenty-fold cross-validation process.

The results show that, statistically, they are equivalent to a random guess (52.1\% correct $\pm 6.2$\%).  Given that the features at each azimuth do not necessarily correspond to those at other azimuths, this comes as no surprise.  It is difficult to distinguish an object given a swath of features related to an azimuth; much easier by applying a Fourier transform on the data, fixing the data in the bins regardless of the azimuth viewed. 

There is nothing exclusive about the Bayesian success with these features; to wit, results from support vector machines on data could be combined with results from Bayesian belief networks to arrive at a classification more accurate than either could provide.

Our results on our FFT-preprocessed, synthesized RCS returns with the kernel adatron that was then processed by the SVM results in the following performance: spheres, 100\%; cylinders, 99.4\%; frustum, 95.3\%; polygons, 95.6\%. These results are comparable to the Bayesian approach with feature extractions. Further, our results are more comprehensive because the individual SVM is classifying the spheres, cylinders, frusta (more complex than cones), and polygons. The computation load with the SVM (including prior feature extraction of FFT) is about 2 MFLOPS.  The computational load with feature extraction and identification by the Bayesian method is 2 GFLOPS~\cite{B:maurerinfo}, a three-orders-of-magnitude greater computational load).  The main problem with the work in~\cite{B:maurer} is the significant preprocessing and feature extraction from the RCS data prior to submitting the information to the Bayesian network.

In conclusion, the kernel adatron, a support vector machine, classifies objects at least as well as the conventional Bayesian network in~\cite{B:maurer} for RCS signature classification, with far fewer computational operations required.  It is especially applicable to time-critical applications where false negatives may have devastating consequences~\cite{B:ruggeri1}\cite{B:ruggeri2}\cite{B:kipersztok}.

\section{Acknowledgments}

This paper has been approved for public release.  Achieving these results came from the contributing work of Peter Nebolsine and Peter Mayer.  They provided the simulated RCS data, several reviews of the results, and suggestions for proceeding forward.  The authors are grateful for their contributions.  The authors also thank the US Army Space \& Missile Command for their support in conducting this research.  We thank Donald Maurer for providing us with feature data, and Merlin Miller, Norman Humer, and Mark Merritt for their assistance.

\end{document}